\documentclass{ieeeaccess}
\usepackage{cite}
\usepackage{amsmath,amssymb,amsfonts}

\usepackage{textcomp}
 \usepackage{balance}
\usepackage{textcomp}
\usepackage{multicol}
\usepackage{multirow}
\usepackage{booktabs} 
\usepackage{url}  
\usepackage{cite}

\usepackage{caption}
\newcommand{\fakeparagraph}[1]{\smallskip\noindent\textbf{#1.}}

 \usepackage{algorithm}
\usepackage{algpseudocode}
\usepackage{lipsum}

\def\BibTeX{{\rm B\kern-.05em{\sc i\kern-.025em b}\kern-.08em
    T\kern-.1667em\lower.7ex\hbox{E}\kern-.125emX}}
    
\newcommand{\nextitem}{\par\hspace*{\labelsep}\textbullet\hspace*{\labelsep}}
\usepackage{graphicx}
\usepackage{resizegather}

\begin{document}

\bstctlcite{IEEEexample:BSTcontrol}

\history{ }
\doi{ }

\title{Multi-Component Optimization and Efficient Deployment of Neural-Networks on Resource-Constrained IoT Hardware}

\author{\uppercase{Bharath Sudharsan}\authorrefmark{1},
\uppercase{Dineshkumar Sundaram}\authorrefmark{2}, 
\uppercase{Pankesh Patel}\authorrefmark{3}, \\
\uppercase{John G. Breslin}\authorrefmark{1}, 
\uppercase{Muhammad Intizar Ali}\authorrefmark{4}, 
\uppercase{Schahram Dustdar}\authorrefmark{5},
\uppercase{\\ Albert Zomaya\authorrefmark{6}, and  Rajiv Ranjan}\authorrefmark{7}}
\address[1]{Confirm SFI Research Centre for Smart Manufacturing, Data Science Institute, NUI Galway, Ireland \\ (e-mail: \{bharath.sudharsan, john.breslin\}@insight-centre.org)}
\address[2]{AVM Solutions UK (email: dinesh.kumar@avmsolutionsuk.com)}
\address[3]{Artificial Intelligence Institute, University of South Carolina, Columbia, USA. (e-mail: ppankesh@mailbox.sc.edu)}
\address[4]{School of Electronic Engineering, Dublin City University, Ireland (e-mail: ali.intizar@dcu.ie)}
\address[5]{Distributed Systems Group, TU Wien, Austria. (e-mail: dustdar@dsg.tuwien.ac.at)}
\address[6]{Albert Zomaya is with the University of Sydney, Sydney, Australia (e-mail: albert.zomaya@sydney.edu.au)}
\address[7]{School of Computing, Newcastle University, Newcastle upon Tyne, UK (e-mail: raj.ranjan@ncl.ac.uk)}
\tfootnote{This publication has emanated from research supported in part by a research grant from Science Foundation Ireland (SFI) under Grant Number SFI/16/RC/3918 (Confirm) and also by a research grant from SFI under Grant Number SFI/12/RC/2289\textunderscore P2 (Insight), with both grants co-funded by the European Regional Development Fund.}

\markboth
{B. Sudharsan \headeretal: Multi-Component Neural-Network Optimization}
{B. Sudharsan \headeretal: Multi-Component Neural-Network Optimization}

\corresp{Corresponding author: Bharath Sudharsan (e-mail: bharath.sudharsan@insight-centre.org).}

\begin{abstract}
 The majority of IoT devices like smartwatches, smart plugs, HVAC controllers, etc., are powered by hardware with a constrained specification (low memory, clock speed and processor) which is insufficient to accommodate and execute large, high-quality models. On such resource-constrained devices, manufacturers still manage to provide attractive functionalities (to boost sales) by following the traditional approach of programming IoT devices/products to collect and transmit data (image, audio, sensor readings, etc.) to their cloud-based ML analytics platforms. For decades, this online approach has been facing issues such as compromised data streams, non-real-time analytics due to latency, bandwidth constraints, costly subscriptions, recent privacy issues raised by users and the GDPR guidelines, etc. In this paper, to enable ultra-fast and accurate AI-based offline analytics on resource-constrained IoT devices, we present an end-to-end multi-component model optimization sequence and open-source its implementation. Researchers and developers can use our optimization sequence to optimize high memory, computation demanding models in multiple aspects in order to produce small size, low latency, low-power consuming models that can comfortably fit and execute on resource-constrained hardware. The experimental results show that our optimization components can produce models that are; (i) 12.06 x times compressed; (ii) 0.13\% to 0.27\% more accurate; (iii) Orders of magnitude faster unit inference at 0.06 ms. Our optimization sequence is generic and can be applied to any state-of-the-art models trained for anomaly detection, predictive maintenance, robotics, voice recognition, and machine vision.
 
\end{abstract}

\begin{keywords}
Edge Intelligence, Neural Networks, Optimization, TinyML, IoT Hardware.
\end{keywords}

\titlepgskip=-15pt

\maketitle

\section{Introduction}\label{sec:introduction}

\begin{figure*}
\centering
  \includegraphics[width=14.5cm]{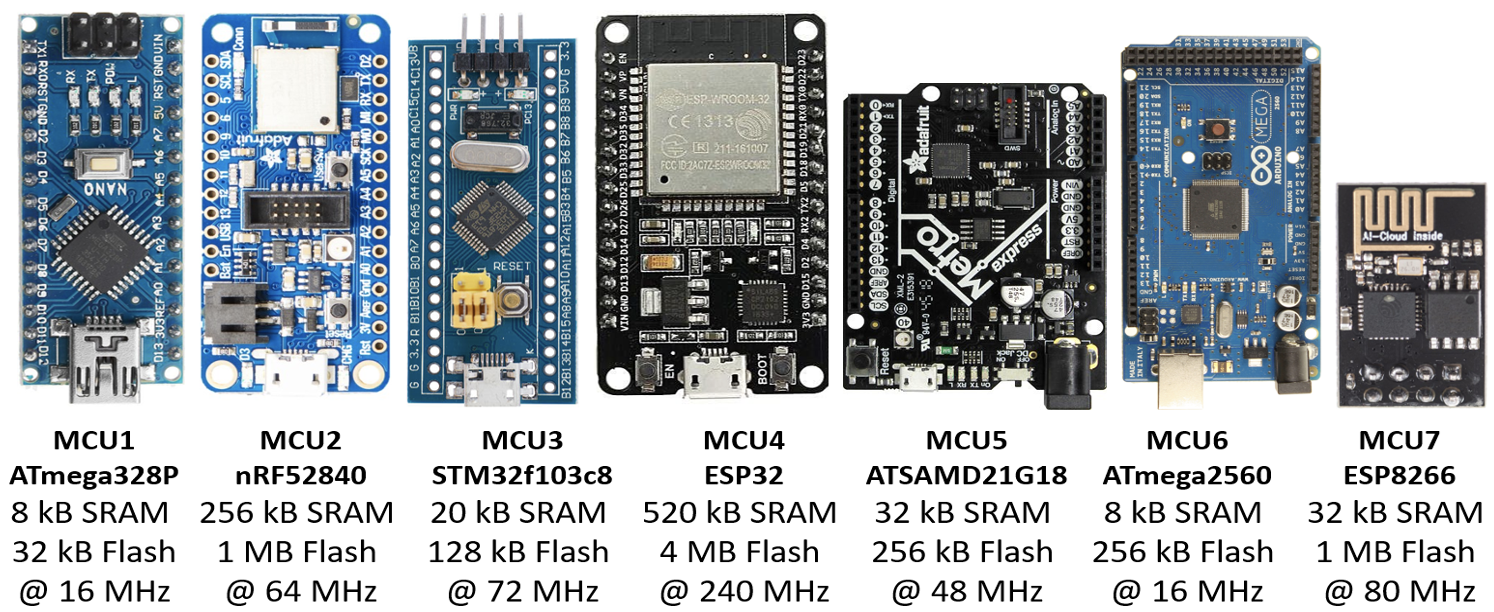}
  \caption{Some of the popular examples of MCU-based hardware that are widely used to design IoT devices.}
  \label{aiot_boards} 
  \vspace{-.5em}
\end{figure*}

\IEEEPARstart{A}{rtificial} Intelligence~(AI) have been used as the principal approach to solve a variety of significant problems in machine translation, video analytics, voice localization, handwriting recognition, etc. Commonly, to provide edge-level AI-functionalities to customers, manufacturers program their IoT devices/products to capture, compress and transmit data (image, audio, sensor readings, etc.) over the network to their central server/cloud where advanced analytics are performed \cite{iot-conf2020}. Although such cloud-based approaches reduce the maintenance cost by keeping the analytics models in one central location, it may not be suitable for most applications \cite{bregar2018improving} because; First, there is a latency caused when transmitting data to a central server for analysis and back to the application. Second, the use of a server for continuous data storage and analysis is expensive because these applications generate high volumes of data. Furthermore, the processing and storage of multiple data streams make the subscription more costly. This design requires a huge amount of reliable bandwidth, which may not always be held for IoT devices \cite{lai_2018_cmsisnn}. Moreover, when employing such an online approach, the hardware cost increases due to the addition of a 4G/WiFi module for communication, which also increases power consumption. Third, even if we assume that we could address latency and bandwidth issues by empowering a sophisticated infrastructure, a large class of IoT applications may not be suitable because of regulations and security concerns of sharing data as there is an involvement of biometric data of residents. For instance, GDPR restricts the sharing of user's private data across organizations. Finally, IoT devices are not self-contained ubiquitous systems since they depend on the internet and cloud services for inferences and model updates \cite{vrevca2020accelerating}.

To address the aforementioned concerns, there is a strong need for an approach which \textit{enable offline analytics on IoT hardware}. Many ML frameworks such as TensorFlow Lite, Caffe2, Apache MXNet, ONNX, etc., exist to deploy Neural Network~(NN) based models on IoT devices~\cite{murshed2019machine}. For instance, Google's TensorFlow (TF) Lite contains a set of tools that help developers to optimize and run TF models (.tflite model file) on IoT devices such as Raspberry Pi, Android and iOS smartphones \cite{sudharsan2021enabling}. However, such frameworks are not suitable for resource-constrained hardware like Micro Controller Units~(MCUs), small CPUs since executing these software frameworks alone requires hundreds of MBs for storage, file system support, high clock speeds, multiple cores, parallel execution units, etc. The majority of IoT devices such as fitness bands, smart plugs, HVAC controllers, etc., are powered by MCUs and small CPUs that are highly resource-constrained. For example, the Arduino Nano is an 8-bit ATmega328 MCU with a 16 MHz clock, 2 kB of SRAM, 32 kB of ISP flash memory, and the NUCLEO-F303K8 is a 32-bit ARM Cortex-M4 MCU  with a 72 MHz clock and 64 kB of flash memory. Fig.~\ref{aiot_boards}, shows some of the popular examples of hardware that are widely used to design IoT devices, and billions of similar specification hardware-based devices have already been deployed in the world.

\begin{table*}[!ht]
\centering 
\scriptsize\addtolength{\tabcolsep}{-3pt}
\begin{tabular}{p{1.1cm}  p{4cm}  p{3.7cm} p{3.3cm}  p{4cm}}
 \toprule\textbf{Approaches}&\textbf{Description}&\textbf{Strengths}&\textbf{Weaknesses}&\textbf{Tools \& Technology} \\ \midrule 

\multirow{1}{*}{\parbox{1.4cm}{ML \\on MCUs}}& Models are trained on data center GPUs, then passed into a hardware and software co-optimization pipeline to obtain resource-friendly models that the edge devices can execute to obtain predictions
  & \nextitem{Reduced inference latency} \nextitem{MCUs are low cost, small form factor, and low power consuming} \nextitem{Wide hardware choices} \nextitem{Minimal risks of memory, app, and hardware crashes} & \nextitem{Optimizers matching challenges. Performance metrics trade-offs} \nextitem{Battery wear, heat \& runtime memory overflow issues} \nextitem{Supports only Embedded C, C++ and Micro Python} & \nextitem{\textbf{Compression:} Quantization, pruning, weight clustering, tensor decomposition, NAS} \nextitem{\textbf{Programming:} Arduino IDE, Atmel Studio, Keil MDK}  \nextitem{\textbf{Conversion:} m2cgen, sklearn-porter, emlearn, TF Micro} 
\\ \midrule
\multirow{1}{*}{\parbox{1.3cm}{AIoT \\ Hardware}} 
& Co-processors/accelerators, edge GPUs, and AIoT boards available to execute models using light version of ML frameworks
& \nextitem{On-board FFT, FPU, KPU, APU support for calculation speedups} \nextitem{Support from well documented and continuously updating ML frameworks like TF Lite, FeatherCNN, OpenNN, Edge-ML}
& \nextitem{Resource management challenges} \nextitem{Model optimization, deep  compression, and neural architecture search efforts} 
 &  \nextitem{\textbf{High-end:} Intel Movidius NCS, Google Coral TPU, NVIDIA Jetson Nano, Intel NUC series, LattePanda, BeagleBone  boards}
 \nextitem{\textbf{Low-cost:} Banana Pi, UDOO BOLT, Rock Pi, Digi Connect core SBC Series, Orange Pi, 96 boards} 
\\ \midrule
\multirow{1}{*}{\parbox{1.5cm}{Cloud-based SaaS}} &  No ML expertise required to perform the SaaS-based model re-training and inference tasks. Users simply need to share image frames of objects or scenes required to be identified. Rest heavy lifting is handled by the subscribed services
 &  \nextitem{Scalability, agility \& rapid development platform}
 \nextitem{Anywhere \& anytime infinite resource on demand}
 \nextitem{Routine and safe data storage} \nextitem{Wide variety of use-case and unstructured data support}
 &  \nextitem{High latency and subscription cost} \nextitem{Privacy concerns} \nextitem{Limited customization and Vendor lock-in} \nextitem{Loss of control}
 & \nextitem{\textbf{Popular:} AWS Rekognition, Google Vision AI, Microsoft Azure CV, IBM Watson Visual Recognition} \nextitem{\textbf{Customizable:} Google AI platform \& Cloud, AutoML, MakeML, IBM Cloud Pak for Data}
\\ \bottomrule													 
\end{tabular}
\newline
\newline
\vspace{-1em}
\caption{Summary of the approaches and hardware used to develop AI-powered IoT devices/products.}
\vspace{-2.2em}
\label{table:sota} 
\end{table*}

\subsection{Motivation}
In the following, we present the research challenges based on our experimental experience and our recent empirical study. They are the core motivation of our research. 

\fakeparagraph{Neural Networks vs Resource-constrained MCUs}
Executing NNs on MCUs-based resource-constrained IoT hardware~(shown in Fig.~\ref{aiot_boards}) is challenging, because  Firstly, the memory footprint~(SRAM, FLASH, and EEPROM) is limited to a few MBs. No secondary memory is added during the design phase of IoT hardware in order to conserve energy and to maintain high instruction execution speeds. On the other hand, NN models routinely contain millions of parameters requiring higher MBs of storage. Secondly, the computation core~(commonly a single ARM Cortex-M CPU) runs only up to a few hundred MHz resulting in low operations per second. Next is the absence of native file system support, no support for floating-point operations, and the inability to perform parallel processing due to the absence of multiple computational units make the execution of NNs more challenging. Finally, for a single inference, such models roughly invoke $10^9$ arithmetic operations and memory accesses, leading to substantial power consumption and heat dissipation, draining the limited battery capacity while testing the device’s thermal limits.

\fakeparagraph{Existing Programming Frameworks} The compression levels and speedups produced by generic optimization toolkits in ML frameworks~(e.g., PyTorch, Tensorflow) are not sufficient since they are targeted for smartphones and better-resourced IoT hardware such as Raspberry Pis, Jetson Nano. The early-stage TF Lite for Microcontrollers \cite{tfmicro} core runtime can fit in 16 KB on an Arm Cortex M3 and run basic NNs on MCUs without needing operating system support or dynamic memory allocation. Here, still, in order to highly reduce the NN size before deploying and executing on MCUs using TF Micro, there is a need to \textit{optimize high memory, computation demanding models in multiple aspects to produce small size, low latency, low-power consuming models that can comfortably fit and execute on resource-constrained hardware presented in Fig.~\ref{aiot_boards}}.

\fakeparagraph{Performance Metrics Trade-offs} During on-board model execution, an IoT application that interacts with the loaded model may demand high performance on a particular metric over others. For example, a real-time IoT device would require ultra-fast inference, while a low-memory device would require the highest model size reduction. So, the challenge is \textit{how to perform optimization that favors particular metrics over others?}

\fakeparagraph{Optimization Compatibility} When a NN model~(optimized using a state-of-the-art method) exceeds the target IoT device hardware's memory capacity by a few bytes margin (a common scenario in practice), there is a need to choose and apply an additional optimization method that is compatible with the previously used optimizer. In such cases, the researchers or developers need to spend days on unproductive work that involves finding a compatible optimizer, then implement it to check if the new compression and accuracy levels are satisfactory. But models cannot be optimized further if they fail to find a method that matches the previous optimizer. So, they either have to tune the model network architecture and re-train from scratch to produce a smaller model (waste of GPU days and electricity) or upgrade the IoT device hardware (loss of money). Hence, in order to speed up the research and development phase (going from idea to product) of AI-powered IoT devices, the researchers and developers need a comprehensive guideline to optimize NN models that can readily be deployed on resource-constrained MCUs-based hardware.

Given the potential of building intelligent IoT applications for offline analytics using NNs, there is a strong need for a mechanism that can optimize models to achieve reduced model size, faster inference, and lower power consumption.

\subsection{Our Approach}

To address the aforementioned challenges and design goals, we propose a multi-component model optimizer that enables the execution of computational intensive NN models on resource-constraint IoT hardware presented in Fig.~\ref{aiot_boards}. The contributions of this paper are as follows:

\fakeparagraph{Multi-component Optimizer Design and Implementation} We propose multi-component model optimizer, a sequence that researchers and developers can follow to optimize various NNs for making it executable on multiple resource-constrained IoT hardware. The proposed design flow is based on a combination of state-of-the-art optimizers. To the best of our knowledge, we are the first to present a complete design flow, with its implementation freely made available on github\footnote{\url{https://github.com/bharathsudharsan/CNN_on_MCU}}. We believe that the transparent design will greatly aid the model optimization steps to be seamlessly integrated into the AI-powered IoT product development life cycle at a minimal cost.

\fakeparagraph{Validation Study and Evaluation Results}  We perform experiments on Convolutional Neural Networks~(CNNs) and show the readers which presented components need to be used together in order to optimize their CNN-based use-case models for; (i) Smallest size (12.06 x times compression); (ii) Best accuracy (0.13\% to 0.27\% improvements); (iii) Ultra-fast unit inference at 0.06 ms (orders of magnitude faster than original models). We also explain how to practically deploy and execute models optimized using our multi-component model optimizer on tiny IoT hardware.

\fakeparagraph{Outline} The rest of the paper is structured as follows:
Section~\ref{sec:sota} presents the state of the art. In Section~\ref{sec:multi}, we present our end-to-end multi-component ML model optimizer. Section~\ref{sec:evaluation} presents the evaluation results and analysis. Section~\ref{concl} concludes our current work and presents the future outlook.

\section{State of the Art}\label{sec:sota}

Table~\ref{table:sota} presents an overview of approaches and hardware centered around the deployment of AI-powered IoT devices/products. It is broadly divided into three categories: \textit{Machine Learning on Microcontrollers}, which focuses on deep optimization of ML models to enable its accommodation and execution on resource-constrained MCUs~(Section~\ref{sec:mlop}); \textit{Artificial Intelligence of Things~(AIoT) hardware}, which presents AI accelerators, GPU-boards and embedded systems that are the target to execute NN models~(Section~\ref{sec:aiot}); \textit{Cloud-based}, which leverage cloud-based software as a service~(Section~\ref{sec:on-cloud}) for inference. In the following, we describe each approach in detail.

\subsection{Machine Learning on Microcontrollers}\label{sec:mlop}
The efforts belonging to this category focus on optimizing ML models to enable their execution on MCUs shown in Fig.~\ref{aiot_boards}. Here, the overall objective is to reduce the model size and execution latency while aiming to maintain acceptable levels of accuracy. 

The MCU hardware platform has recently become an attractive target to run models due to TensorFlow Micro and MCU-targetted optimized kernels from CMSIS \cite{lai_2018_cmsisnn}. In this domain of enabling intelligence on resource-constrained devices, the authors in \cite{lai2018cmsis} have implemented a tree-based algorithm, called Bonsai, for efficient prediction on IoT devices. High accuracy predictions were obtained in milliseconds even on slow MCUs and were able to fit in a kB of memory. Similarly, ProtoNN, a k-Nearest Neighbor (KNN) inspired algorithm with several orders of lower storage and prediction complexity was proposed in \cite{gupta2017protonn} to address the problem of real-time and accurate prediction on resource-scarce devices. Both \cite{lai2018cmsis} and \cite{gupta2017protonn} are tailored prediction algorithms that can fit in resource-scarce devices and show superior performance. Such techniques are lightweight and show low execution latency, making them well-suited for offline analytics. However, the design flow used by then cannot be applied for NNs. In the remainder of this section, we present NN optimization techniques.

In the domain of NN optimization, various techniques are applied to enable the deployment of NNs on IoT devices. For instance, \textbf{Model design} techniques emphasize on designing models with a reduced number of parameters without compromising accuracy, thereby enabling it to fit and execute within the available IoT device memory~\cite{deeplearning-edge}. Also, \textbf{Model compression} techniques such as quantization \cite{chowdhury2022deepqgho} and pruning~\cite{deeplearning-edge} can be used. Where quantization takes out the expensive floating-point operations by reducing it to a Q-bit fixed-point number, and pruning removes the unnecessary connections between the model layers. Google’s TensorFlow Hub offers pre-trained and optimized versions of popular models such as Inception, Xception, MobileNets, Tiny-YOLO, ResNet, EfficientNet, etc. that can identify hundreds of classes of objects, including people and animals. For example, in \cite{sudharsan2019ai}, the Mobilenet-SSD is used to enable their camera-based Alexa smart speaker prototype to detect and identify objects in the room (Alexa custom skill), and in \cite{aics2019} a Deep Neural Network based biometric authentication was used to address the cybersecurity risks by smart speaker users. Such pre-optimized models can be readily loaded only on better-resourced IoT devices like Raspberry Pis, Coral boards, Jetson Nano, etc., not on resource-constrained MCUs, small CPUs, which lack even the basic file-system support.

\subsection{Artificial Intelligence of Things~(AIoT) Hardware}\label{sec:aiot}

While the model compression/shrinking techniques can help NN models to run on IoT devices, it is still a challenge to deploy dense NN models on MCUs and obtain real-time inference results \cite{vrevca2020accelerating}. To address this challenge, the NN workloads are offloaded from an IoT device to powerful edge accelerators/co-processors \cite{sekanina2021neural}. In such a scenario, an IoT device re-directs the data stream to accelerator hardware, where it processes the data and sends back the corresponding inference results. In today's IoT hardware market, there is a new category of IoT edge hardware emerging named Artificial Intelligence of Things~(AIoT) devices, which can be divided into two categories, as mentioned below:

\begin{figure*}
\centering 
  \includegraphics[width=\textwidth]{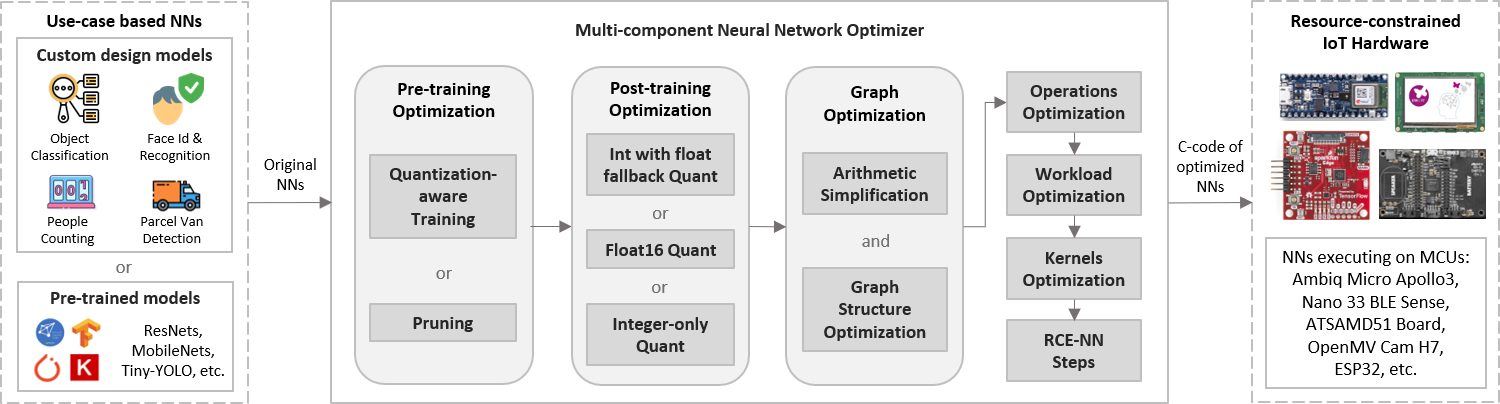}
  \caption{The architecture of our multi-component model optimizer: Sequence to follow for optimizing Neural Networks to enable its execution on resource-constrained AIoT boards, small CPUs, MCUs based IoT devices.}
  \label{optimizer_arch} 
  \vspace{-1em}
\end{figure*}

\fakeparagraph{AI Accelerators and Dedicated GPU Boards} The winning model of the ImageNet challenge showed 11.8\% improvements in classification accuracy. i.e., from 84.7\% in 2012 (winner AlexNet) to 96.5\% in 2015 (winner ResNet-152). Such exceptional accuracy improvement comes with high computational complexity. For instance, 1.4GOPS (Giga Operations Per Second) is required by AlexNet, while ResNet-152 consumes 22.6GOPS for the same task of processing a single 224×224 image. In our resource-constrained scenario, such high demand for hardware resources is prohibitive. Next is energy consumption. For instance, just for DRAM accesses, running a 1-billion connection NN at 30Hz would approximately require 30Hz × 1G× 640pJ = 19.2W, which is again prohibitive in our case. For such challenges, custom hardware accelerators \cite{5981829, 5272559} (tailored design based on the computation pattern of the use-case model) are available specifically to offload the NN workload to achieve higher efficiency than GPUs. These designs reduce the memory access expense, and second memory transfer and data movement are optimized. Such accelerators treat the models as black boxes.

For generic design and development scenarios, to speed up inference or on-device training during active learning scenarios, hardware manufacturers have developed low power and cost GPU plugins to improve parallel processing performance. For instance, Intel's Movidius Neural Compute Stick 2~(NCS) and Google's Coral Tensor Processing Unit~(TPU) USB accelerator are some of the co-processors that can be plugged into the USB port of edge device like Raspberry Pis, BeagleBones, etc. to accelerate machine vision tasks such as, drone-based 3D mapping, contextual awareness, crowd counting, checking compliance of the use of face masks, etc. There are times when CPU-based devices with co-processors/accelerators may not be enough for running DL models. To address this issue, hardware manufacturers offer GPU-based dedicated development boards, which tend to have better ML framework compatibility and NN execution performance. NVIDIA's Jetson Nano and Google Coral development board are popular examples.

   \begin{figure*}
\centering
  \includegraphics[width=13cm]{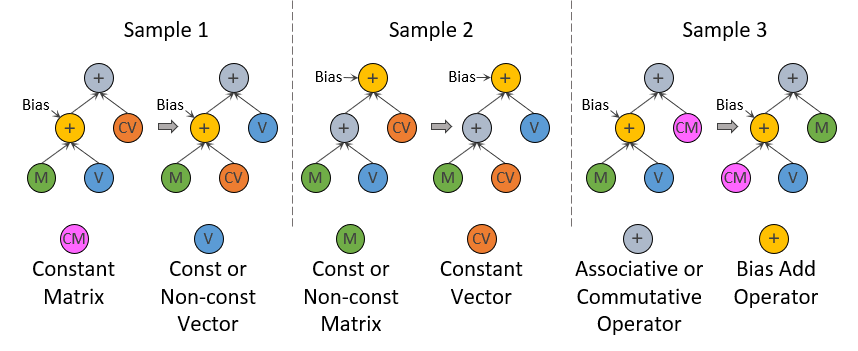}
  \caption{Model graph optimization by arithmetic simplification: A few example known inputs-based arithmetic re-writes to simplify the graph.}
    \vspace{-1em}
  \label{known_inputs} 
\end{figure*}

In this paper, we significantly compress and simplify the models before offloading it to the hardware accelerators, in order to reduce workload redundancy, save memory bandwidth and also computation cycles.

\fakeparagraph{Resource-Constrained Embedded Systems} The devices belonging to this category are embedded systems designed using a variety of low hardware specification chipsets, which are usually MCUs and small CPUs \cite{sudharsan2021enabling, bregar2018improving}. The nRF52840 Adafruit Feather, STM32f103c8 Blue Pill, ATSAMD21G18 Adafruit METRO, etc., are the embedded boards that are widely used during the design phase of IoT devices, and billions of similar specification hardware-based devices have already been deployed in the world~(e.g., smart bulbs, smart plugs, HVAC controllers). There is another set of better-resourced IoT hardware named AIoT boards. The ESP32-CAM, Sipeed MAIX Bit, M5 StickV AI Camera, Sipeed Maix Amigo, OpenMV Cam H7 are popular AIoT boards that are purpose-built to deliver high performance in a small physical and power footprint, enabling users to deploy high-accuracy AI at the edge. The traditional small CPUs and MCUs based IoT hardware from the price category of AIoT boards does not have a camera module and also lack inbuilt hardware accelerators such as APU (Accelerated Processing Unit), KPU~(convolution operation accelerator), FPU~(Floating-point accelerator), and FFT~(Fourier transform accelerator). The competitive price range and its AI-friendly hardware specification make it a suitable choice to program and use as privacy-preserving offline analytics performing edge device. 

Fig. \ref{aiot_boards}, shows a few popular MCU-based development boards with their specifications. These open-source MCU boards and the above described AIoT boards are $\approx$ 10 × cheaper, $\approx$ 3× smaller, and $\approx$ 12 x less power consuming than the AI accelerators and dedicated edge GPU boards mentioned in the previous section. This is because such boards are powered by single-chip processors that are very cheap~(few \$), tiny~($\approx$ 1 $cm^{2}$), and highly energy-efficient~($\approx$ 1 mW). In this paper, we present how to optimize NN models in multiple aspects to obtain a small size, low latency and power-consuming model that can readily be deployed on resource-constrained embedded systems and AIoT boards.

\subsection{Cloud-based Software as a Service~(SaaS)}\label{sec:on-cloud}

This approach leverages cloud-based services. Various cloud vendors~(Such as Amazon, Google, Microsoft) have developed ML-based analytics platforms. A typical approach is where a device sends a data stream over the network to a cloud-based platform, where all the processing takes place and analytics results are sent back to the IoT device~\cite{9097404}.

For example, our previous work~\cite{iot-conf2020, iot-conf2020-demo} presents an implementation design that uses AWS as a base technology. We choose AWS Rekognition service, which lets developers develop several NN-based computer vision capabilities on top of scalable and reliable Amazon infrastructure. Rekognition offers services, which can be divided into two categories: First, the developer can leverage pre-trained algorithms (prepared by Amazon) to identify objects, people, text, scenes, and activities in videos, as well as detect any inappropriate content. Second, Rekognition Custom labels enable the developers to build use case specific NN-based video analytics capabilities to detect unique objects and scenes. 

\section{Multi-component Optimizer Design}\label{sec:multi}

Fig. \ref{optimizer_arch}, shows the proposed end-to-end multi-component model optimizer. This optimizer takes a Neural Network (NN) model as an input and produces a highly optimized version of the input NN that can run on low resource, cost and power IoT hardware such as MCUs~(shown in Fig. \ref{aiot_boards}). In each of the following sections, we present different phases of the proposed multi-component optimized design.

\subsection{Pre-training Model Optimization} \label{pretrain}

This section presents pre-training techniques to optimize NN models.

\subsubsection{Pruning}

Model pruning induces sparsity in a NN's various connection matrices to reduce the number of non-zero parameters. The concept of pruning models enables trading off a small portion of a model's quality for reduced model size, lesser inference time, and improved thermal and energy efficiency.

The pruning technique we implemented as a part of our multi-component optimizer is from \cite{zhu_2017_to}. This component prunes the model's network connections during the training step, to reduce the size and working memory with only a minimal impact on model accuracy. In this pruning method we implemented, we initially increase the sparsity from an initial value $s_{initial}$ starting from 0 to a final sparsity value $s_{final}$, for $k$ pruning steps. The training step starts at $t_0$, with a pruning frequency of $\delta t$.
\begin{equation} \label{pru}
s_{t} = s_{final}+(s_{initial}-s_{final}) \left(1-\dfrac{ t-t_0}{n\delta t}\right)  \text{ for }t  \in {t_0...t_0+k \delta t}
\end{equation}

Binary weight masks are variables that take the same size \& shape of the model layer’s weights. Every layer that we chose to prune should contain this mask since it is used to determine which weights engage in the forward execution of the graph. We update these masks, every $\delta t$ steps during the training, to gradually increase the sparsity of the network. Once the model achieves the target sparsity $s_{final}$, we stop updating the masks. We use Eqn. (\ref{pru}) to remove redundant connections, while gradually reducing the prune count of weights in every iteration, since the number of weights in the network keeps reducing. We recommend the user's to realize this method for achieving model sparsity. The highlights of this method are; it is independent of the model's network property \& its constituent layers, hence users can apply it to a wide range of models (can be used by models without structural properties, e.g. LSTMs). Not much hyperparameter tuning is needed (users don't have to select slopes and weight-threshold), and it performs well across various models. 

\subsubsection{Quantization-aware Training}
This section briefly explains how to perform quantization-aware training~\cite{ krishnamoorthi_2018_quantizing}, of a model which is aimed to  execute on tiny hardware. Here, this technique first considers quantized weights in full precision representation to simulate and inject quantization error into training, thus enabling the weights to be optimized against quantization errors. This quantization error is modelled using fake quantization nodes, which simulates the effect of quantization in both forward and backward passes. These fake quantization operations are added to all required locations in the model by rewriting the training graph (by creating a fake quantized training graph). 

Based on our evaluation results in Section~\ref{sec:evaluation}, this method significantly improves the latency-vs-accuracy trade off for the majority of use-cases. In cases when performance does not improve, then the user should directly train the model and perform post-training quantization using any one method we provided at Section \ref{PTQ} since it is broadly applicable to all models and does not require training data during quantization.

\subsection{Post-training Model Optimization} \label{PTQ}
The latest versions of widely used edge-friendly NN models need additional optimizations to fit and run comfortably within MCUs-based devices. This section presents methods that can be applied to pre-trained models (such as Inception, Xception, Mobilenet, Tiny-YOLO, Resnet, etc.) as well as custom-designed ones. The optimization methods that are our post-training model optimization components quantize the models by reducing the precision of its weights to save memory and simplify calculations often without much impact on accuracy. The optimization methods that are our post-training model optimization components quantize the models by reducing the precision of its weights to save memory and simplify calculations often without much impacting the accuracy of a model.

\begin{figure*}
\centering
  \includegraphics[width=12.9cm]{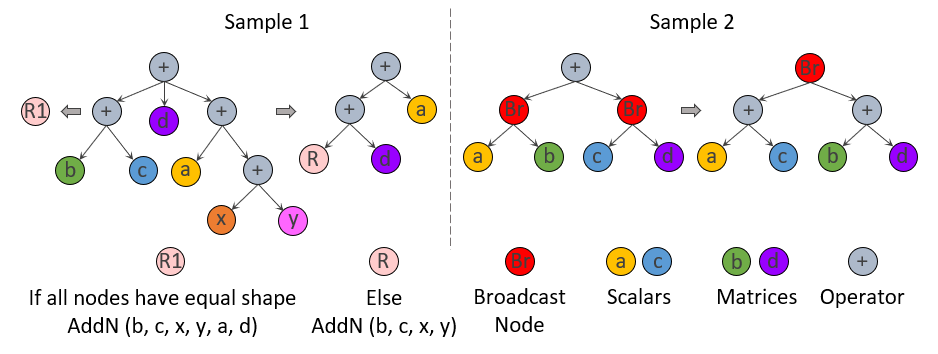}
  \caption{Model graph optimization for spread/broadcast minimization: Re-ordering of operator and input nodes to reduce the graph execution cost.}
  \label{spread_min} 
  \vspace{-1em}
\end{figure*}

\fakeparagraph{Weights only Quantization}
This technique reduces the weight's precision in the network from float to 8-bits precision. Our evaluation results (presented in Section~\ref{sec:evaluation}) shows that the model size is significantly reduced. We adopted the state-of-the approach \cite{wu2020integer} to quantize NN weight $w$ to a Q-bit fixed-point number $quant(w)$, by using the quantization function~(Eqn. (\ref{eqnone})).
\begin{equation} \label{eqnone}
quant(w) = clip_{[-1,1)}(2^{-(Q-1)}.round(w.2^{(Q-1)})
\end{equation}
Here $clip_{[-1,1)}(x) = max(a,min(x, b))$, and the corresponding INT-Q representation of CNN's weights is $W = quant(w).2^{(Q-1)}$. The same Eqn. (\ref{eqnone}) is applied to any activation values as well. It converts the weights and activations of a model to Int-8 data type since they are the most natural type to fit in 32-bit MCU registers.

\fakeparagraph{Weights and Activations Quantization} 
Similar to the weight quantization method presented earlier, activations alone can also be quantized to 8-bits with almost no accuracy loss. We studied multiple existing techniques and found out that Symmetric per-channel, Asymmetric per-layer, and Asymmetric per-channel are well-suited techniques to quantize both weights and activations~\cite{krishnamoorthi_2018_quantizing}. 

We  select per-channel quantization with asymmetric ranges over other techniques since it provides close to floating-point accuracy for a wide range of networks~\cite{krishnamoorthi_2018_quantizing}. This method quantizes both the weights and activations to INT-8 values. Hence, the convolution in NNs takes the following form
\begin{equation} \label{eqntwo}
     \psi(w,x) = 2^{-2(Q-1)} \sum_{i \epsilon D}W_{i}X_{i} \doteq 2^{-2(Q-1)} . \phi(W,X) 
\end{equation}
In Eqn~(\ref{eqntwo}), $D$ is the number of input channels, $\psi$ is NN's convolution operation, and  $\phi(W, X)$ is an accumulator containing high precision values, in our case, Int-32 for Int-8 operands. We recommend this Int-8 quantization method since it outperforms the Fine-Grained Quantization (FGQ) method (2 bits for weight quantization) and Incremental Network Quantization~(INQ) method (5-bit weight floating-point activation) by preserving accuracy while also providing run-time improvements. Next, we quantize the original NN's Float32 weights and activations to Float16 values. Users can use this Float16 quantization when they want to achieve reasonable compression rates (we obtain approx. 6x compression), without loss of precision (we experience only 0.01 \% loss in accuracy). Also, Float16 models run on small CPUs without modification.

\fakeparagraph{Joint Pre and Post-training Model Optimization} If users want to achieve more than 11x size reduction, let's assume when they aim to execute Inception v3 (23.9 MB after post-training quantization) on a AIoT board, which only 16 MB Flash memory, we recommend performing \emph{joint model size optimization}. Here, first, any of the pre-training optimization methods from Section \ref{pretrain} has to be applied to the model, followed by its Int-8 post-training quantization using the technique we provided in Section \ref{PTQ}.

\subsection{Graph Optimization} \label{GO}
The interior of trained models is a  graph with defined data flow patterns \cite{asif2021graph, chen2018tvm}. This graph contains an arrangement of nodes and edges, where the nodes represent the operations of a model, and graph edges represent the flow of data between the nodes. We target the graph level since it is backend independent, interoperable, applicable to both offline and at runtime execution of a NN model. 

This section presents techniques~(optimizers), which can be leveraged to optimize graphs of NN to improve the computational performance of NN while reducing peak SRAM (memory) usage on MCUs, thus enabling the execution of larger models on tiny memory footprints. In the following, we present graph optimization in sequential steps.

   \begin{figure*}
\centering
  \includegraphics[width=12.9cm]{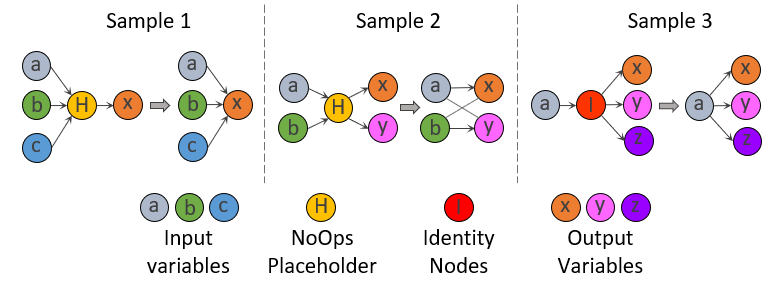}
  \caption{Model graph structure optimization: A few example re-writes of the graph after removing the NoOps and Identity nodes.}
  \label{structure_opt} 
  \vspace{-1em}
\end{figure*}

\subsubsection{Arithmetic Simplification} 
To improve the graph's performance, we propose to simplify its arithmetic by performing the tasks we mention below.

\fakeparagraph{Re-write arithmetic based on known inputs}
Arithmetic re-writes rely on known inputs. As shown in Fig. \ref{known_inputs}, the known constant vector is grouped with an unknown vector that might be constant or non-constant.  After performing such re-writes, if the unknown vector turns out to be a constant, then graph performance improves. In Eqns~(\ref{go1})--(\ref{go3}), we provide a few more examples. 
\begin{equation} \label{go1}
Sub(c_0, Sub(x, c_1)) \quad \textrm{re-written as} \quad Sub(x, c_0 + c_1) 
\end{equation}
\begin{equation} \label{go2}
Conv2D(c_0 * x, c_1) \quad \textrm{as} \quad Conv2D(x, c_0 * c_1)  
\end{equation}
\begin{equation} \label{go3}
\begin{split}
Concat([x, c_0, c_1, y]) \quad \textrm{as} \\ \quad Concat([x, Concat([c_0, c1]), y) 
\end{split}
\end{equation}
\fakeparagraph{Trivial operations removal} We propose to identify and remove transpose, reshape, reverse, and shuffle operations on 1D vectors/tensors. Followed by simplifying operations such as squeeze (removes dimensions that have elements with size one), pad (pads tensors), tile (generates tensor that is input replicated by n times), and slice (extracts a slice of from input) by replacing them with their respective identity operations.

\fakeparagraph{Flattening operations} As shown in the Eqn~\ref{flat-1}, we simplify arithmetic by performing flattening operations.
\begin{equation} \label{flat-1}
w + x + y + z \quad \textrm{as} \quad AddN(w, x, y, z) 
\end{equation}
In this same step, we also perform aggregation simplification of nodes, where we remove nodes that have only a single input and no control dependency. As shown in the Eqn~\ref{flat-2}, in this task we rewrite the aggregation operations that have more than two identical inputs, in order to eliminate multiple sum operations. 
\begin{equation} \label{flat-2}
AddN(a, a, a, ..., a) \quad \textrm{as} \quad Mul(const(N), a) 
\end{equation}
\fakeparagraph{Hoisting} Bulky loops when executed many times results in a significant consumption of resources. In order to simplify the loops, in this task, we first propose to pull out loop-invariant sub-graphs away from loops. For example, in the Eqn~\ref{fig:hoist1}, we pull out the variable $a$, which results in replacing three $*$ operations with one, but introduces three $+$ operations which is less costly than a $*$ operation. Hence, when this simplified Eqn is run in a loop, conservation can be achieved in every iteration.
\begin{equation}\label{fig:hoist1}
 AddN(a * x, y * a, a* z) \quad \textrm{as} \quad a * AddN(x+y+z)  
\end{equation}
In this same task, we next propose to hoist chained unary operations that are nested inside operators. We show the Eqns (\ref{fig:hoist2}) - (\ref{fig:hoist3}) as examples of this hosting task. 
\begin{equation} \label{fig:hoist2}
\begin{split}
Exp(Sin(a)) \text{, for a in Split}(b) \\ \quad \textrm{as}  \quad  Split(Exp(Sin(b), a)  
\end{split}
\end{equation}
\begin{equation} \label{fig:hoist3}
\begin{split}
Concat([Exp(Sin(a)), Exp(Sin(b)), \\Exp(Sin(c))]) \quad \textrm{as} \quad  Exp(Sin(Concat([a, b, c])))  
\end{split}
\end{equation}
\fakeparagraph{Simplification by reducing node counts} Each node in the graph of a NN model represents an operation such as Conv2D, MatMul, Add, etc. Here we propose to perform rewrites that simplify the graph by reducing the number of nodes that in turn reduces the number of required operations. For example, in Eqn~\ref{simp-1}, we simplify by reducing the three numeric operators $+$ nodes to one $*$ node, and two logical operators $!$ and $>$ nodes into one $<=$ node.
\begin{equation} \label{simp-1}
  a + a + a  \quad \textrm{as} \quad  3 * a.  \quad !(a > b)  \quad \textrm{as} \quad  a <= b 
\end{equation}
In this same task, we leverage the multiplication and division operator's distributive and commutative properties in order to take out common factors/denominators. To present this concept, we take the Eqns of the graph patterns that frequently occur when regularizing gradients during model training. Here, in Eqn~\ref{go12}, we take out the common factor from the aggregate nodes where all input nodes are Mul. Similarly, in Eqn~\ref{go13}, we take out a common denominator.
\begin{equation} \label{go12}
\begin{split}
 AddN(Mul(a, x_1), Mul(a, x_2), \dots, Mul(a, x_n)) \\ \quad \textrm{as} \quad  Mul(a, AddN(x_1, x_2, \dots, x_n)) 
\end{split}
\end{equation}
\begin{equation} \label{go13}
\begin{split}
 AddN(Div(a, x_!), Div(a, x_2), \dots, Div(x_n, a)) \\ \quad \textrm{as} \quad  Div(AddN(x_1, x_2, \dots, x_n), a)
\end{split}
\end{equation}
\fakeparagraph{Spread minimization} Shapes of two arrays are compatible only when each of their dimension pair is equal. Broadcasting is the method to make arrays have compatible shapes so they can be used during arithmetic operations. Here, we propose to group similar data types to \emph{minimize broadcast}. For example, in below Eqn~\ref{spread}, we separate arrays from scalar values, then group similar types of data. The resultant simplified Eqn~\ref{spread} is less costly to execute since performing operations between the same type of data is faster and simpler.
\begin{equation}\label{spread} 
\begin{split}
   (a[x] + scalar0) + (b[x] + scalar1) \\ \textrm{as} 
   \quad (a[x] + b[x]) + (scalar0 + scalar1)   
\end{split}
\end{equation}
To achieve higher levels of computational cost reductions, we provide advanced spread/broadcast minimization tasks that rewrites a group of binary associative operators (Add or Mul) and also reorder inputs as shown in Fig. \ref{spread_min}. In this same task we recommend users to bypass or remove redundant reshape and broadcast nodes when the shape of the input matches the target shape.

\subsubsection{Graph Structure Optimization}

We propose the below listed tasks that when realized will optimize the graph for efficiency. 

\begin{itemize}
    
    \item  Remove loop-invariant sub-graphs. i.e., remove the loops that are true both before and after iterations.   
    
    \item  Remove dead branches/ends from the graphs. So, during execution on MCUs, backtracking from the dead-end is not required to progress in the graph.
    
    \item  Find loop-trip counts (number of times a loop executes), then remove loops with zero trip-counts, and remove control flow nodes when the trip-count is one.
    
    \item  Replace recurrent subgraphs with optimized kernels/executable modules. For example, when a recurrent subgraph containing Conv2D + FusedBatchNorm + BiasAdd has to be executed, during runtime, we propose to dynamically load and unload the kernel/executable module (stored in EEPROM of MCUs) as a more efficient replacement for this subgraph.
    
    \item  Use a transitive reduction algorithm on the entire graph to remove redundant control edges. Then, shorten the critical path of a model step by rearranging control dependencies. 
    
    \item This task when realized, reduces the graph's size resulting in processing speedups. Here, we identify and remove nodes that are effectively NoOps (a placeholder of control edges) and Identity nodes (outputs data with the same content and shape of input). We perform this removal only if the product of the number of inputs and the number of outputs is less than or equal to the sum of the number of inputs and the number of outputs. In Fig. \ref{structure_opt}, we show this briefed method.
    
    
    \item Even just removing nodes such as stop gradient nodes (not useful during execution, since its scope ends after building the graph) and Identity ops show speedups on MCUs. When performing aggressive optimization, we recommend not to remove any of the control dependency nodes since that might end up creating many additional edges. Also, do not alter the nodes connected to functions as doing so might cause run time errors. Next, although removing nodes driven by peripherals or other devices is beneficial, yet from our experience, we recommend not to remove them since they can save us more on the sensor to device communication cost. Similarly, we recommend not removing the reference values receiving/saving nodes as doing so converts references to non-references and will cause un-spottable issues since, in the graph partitions, these non-reference values will not be shared more than once. Hence such nodes need to be preserved in order to store the once shared non-references values.

\end{itemize}

\begin{figure*}
\centering
  \includegraphics[width= 12 cm]{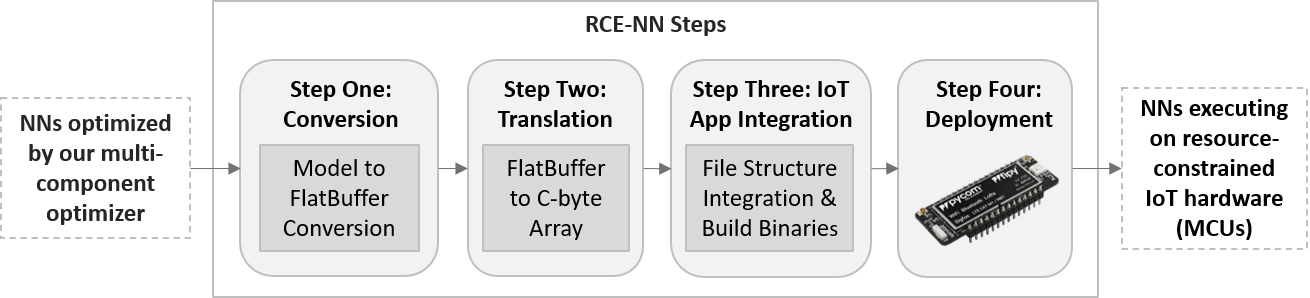}
  \caption{RCE-NN steps to deploy and execute optimized models on resource-constrained IoT hardware.}
  \label{fig:rce-nn_steps} 
  \vspace{-1em}
\end{figure*}

The time consumed by the presented graph optimization component to produce the optimized version of the original model depends on complexity $\mathcal{T_c}\left(|O| 2^{|O|}\right)$, where $|O|$ is the total operators count. Since the latest network architectures contain hundreds of operators, our proposed component is best suited to run on better-to-high resource devices such as standard GPUs or at least laptop CPUs. But our component-generated optimized models can be comfortably executed by the inference software on any IoT hardware.

\fakeparagraph{Joint Graph and Post-Training Model Optimization} In the previous section, we presented our standalone model graph optimization component that can be applied during the model training phase and also suitable for any pre-trained marketplace models. To perform joint model optimization, the two-steps graph optimization component from Section \ref{GO} needs to be applied to the original un-optimized CNN, followed by using any of the post-training model optimizers from Section \ref{PTQ}.

\subsection{Operations Optimization} \label{OO}
When designing ML models aimed to execute as an application on low-specification IoT hardware, only a limited subset of operations can be used in order to keep the operational cost low \cite{chen2018tvm}. In this section, we explain the operations optimization technique that is a part of our multi-component model optimizer.

While designing this component, we viewed CNN operations as absolute arithmetics, aiming to execute it without system-specific functionalities and dependencies on \emph{bare metal} resource-constrained AIoT boards, MCU chips, small CPUs, all of which lack file systems and OS support. We notice that more than 90\% arithmetic operations are used by convolutional~(CONV) layers, so we already convert floating-point operations into int-8 (fixed point) in Section~\ref{PTQ}, which resulted in model size reduction and improved inference performance. Next, taking inspiration from \cite{chollet_2016_xception}, we decompose (depthwise separation) the 2-D CONVs, followed by 1-D CONVs, aiming to reduce parameters and operations count, enabling denser \& deeper CNN architectures executable on low-specification hardware. E.g. a 2D-separable filter $ \phi_{sep}$ has a unary rank, $ \phi_{sep} = 1$. This can be re-written/replaced with two 1D filters, $ \phi_{Ax1}$ \& $ \phi_{1xB}$. When using this depth-separation concept on 3D filters, a regular 3D convolution uses $C * A * B$ multiplications, whereas a depth-separable 3D convolution only requires $C + A + B$ multiplications. In a few CNNs, we were not able to separate the filters. So in such situations, we recommend the users to forcefully separate the high/full-rank filters by penalizing it during training \cite{sironi_2015_learning}. We also recommend another alternative, post-training separation approach where we approximate the layer's weights into smaller sets of $n$ low-rank filters. If done so, only $n * (C+A+B)$ multiplications will be required to execute one 3D-convolution.

\fakeparagraph{Joint Operations and Post-Training Optimization} Even if the models such as MobileNet and SqueezeNet are manually designed to execute within a tight memory budget, it would exceed the AIoT board's capacity by over 5 x times. Hence, we propose to first optimize the operations of any model using the technique from Section \ref{OO}, then apply any of the post-training model optimizers we provide in Section \ref{PTQ}.

\subsection{Workload Optimization} \label{work_opt}

The complexity and size of the model have an impact on the workload. Larger and denser models lead to increased processor workload and result in a higher duty cycle. In such cases, when users execute models with multiple dense layers, the hardware executing the models spend more time working and less time idle resulting in elevated power consumption and heat output. In the following, we present techniques that can be used to reduce workloads. The methods we recommend apply globally, i.e., not biased towards local performance optimizations for a single operation, as in many previous works.

\begin{figure*}
\centering
  \includegraphics[width=14cm, scale = 1]{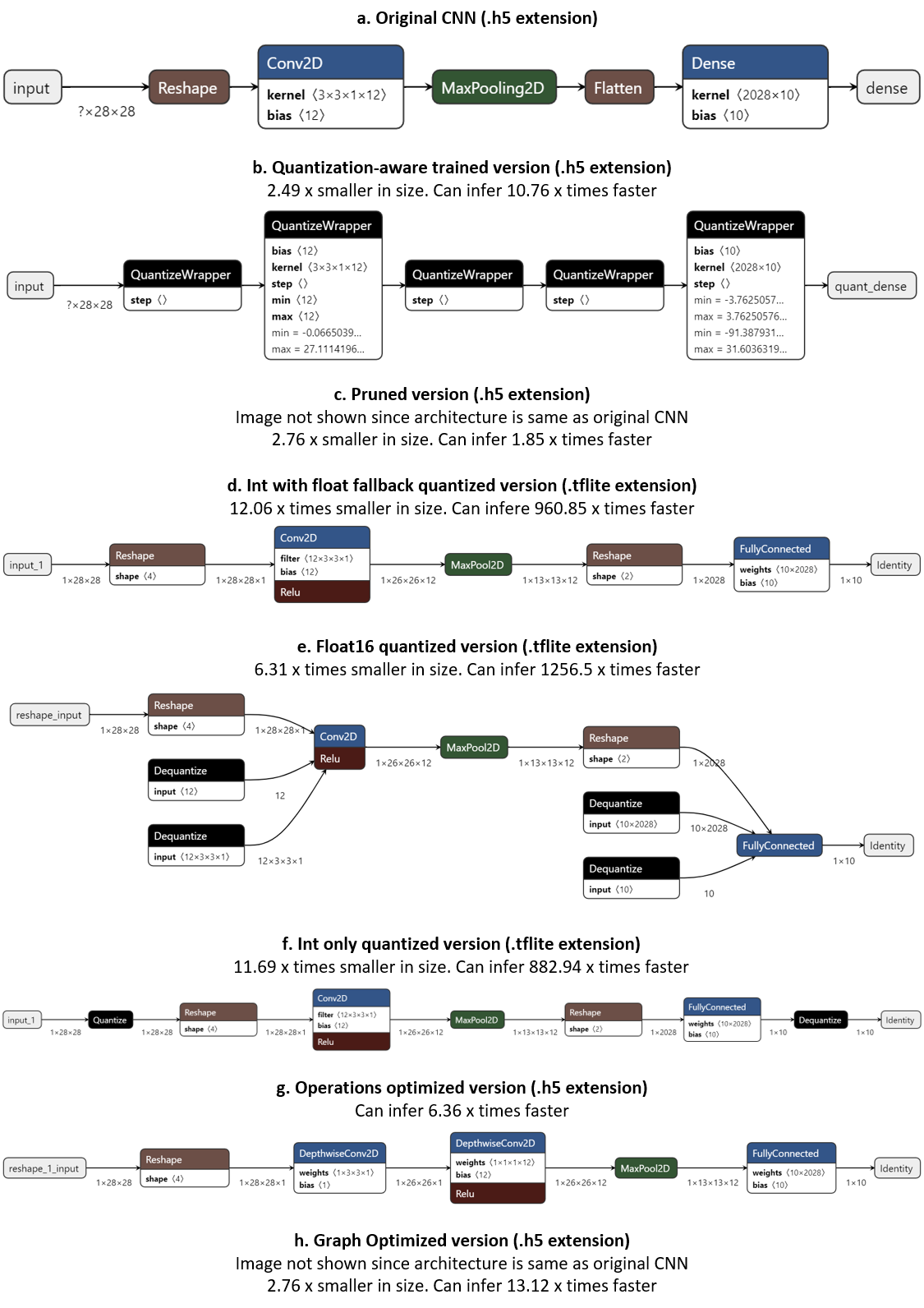}
    \caption{Visualizing and comparing the size and inference time of a CNN with its pre-training, post-training, operations, and graph optimized versions: The a. Original CNN is compared with its b. Quantization-aware trained version, c. Pruned version, d. Int with float fallback quantized version, e. Float16 quantized version, f. Int only quantized version, g. Operations optimized version, h. Graph optimized version.}
  \label{model_comp} 
\end{figure*}

\fakeparagraph{Input data reduction} when the sampling rates of sensors are high, a unit inference result cannot be produced by the MCUs using the acquired data within a sampling interval. In such scenarios, computationally inexpensive low pass filters~\cite{dawit2019edge} should be used to reduce the volume of data, which improves the quality of the data (reduced noise), allows a sufficient time interval to perform the inferencing, and also reduces workload at the same time. 


\fakeparagraph{Hardware accelerators} To improve onboard NN workload execution efficiency, hardware accelerators (e.g., co-processing units) can be used to offload the bulk of NN workloads (convolutional layers) to nearby accelerators. Nowadays, hardware-based NN accelerators are used to bring AI capabilities to better-resourced edge devices like Raspberry Pi, BeagleBoard, etc. which can already run light versions of ML frameworks. In our scenario, to achieve a small form-factor and to keep the hardware cost ultra-low, we chose the more resource-constrained devices~(Fig. \ref{aiot_boards}), which are not capable of utilizing off-the-shelf accelerators~(required software that is larger than the entire available memory) \cite{sekanina2021neural}.

For successful offloads, we recommend storing the C code of the optimized model to be executed in a shared memory location~(EEPROM) that can be accessed via common load/store ports. This accelerated convolution approach can readily be mixed with convolution offload demands (during inference) from other threads running in the same MCU core or on co-MCUs/processors. This parallel offloading approach leads to internal data reuse, hence improving inference performance and the energy efficiency of the AIoT edge devices. In cases where users are ending up with inefficient workload transfers that void the mentioned benefits, we recommend them to offload the processing to the inbuilt KPU, FFT units.

\begin{table*}
\centering
\small\addtolength{\tabcolsep}{-3pt}
\caption{Comparing original models with its: a. Pre-training optimized model versions, and b. Post-training optimized model versions. \\ Changes in CNN's. i. Size. ii. Test Accuracy. iii. Unit inference time.  iv. Model load \& test set inference time.}
\label{pre_and_post}
\begin{tabular}{|c|l|c|c|c|c|c|} 
\hline
\multicolumn{2}{|c|}{\begin{tabular}[c]{@{}c@{}}Original \\ models \end{tabular}}                                                                 & \begin{tabular}[c]{@{}c@{}}a. Q-aware\\ trained version \end{tabular}            & \begin{tabular}[c]{@{}c@{}}a. Pruned\\ version \end{tabular}                     & \begin{tabular}[c]{@{}c@{}}b. Int with float fallback \\ quantized version \end{tabular} & \begin{tabular}[c]{@{}c@{}}b. Float16 \\ quantized version \end{tabular}       & \begin{tabular}[c]{@{}c@{}}b. Int only\\ quantized version \end{tabular}        \\ 
\hline
$CNN_1$ & \begin{tabular}[c]{@{}l@{}}i. 271.4\textcolor[rgb]{0.129,0.129,0.129}{~}KB\\ ii. 89.19 \%\\ iii. 326.69 ms\\ iv. 1419.27 ms \end{tabular} & \begin{tabular}[c]{@{}c@{}}108.7 KB\\88.32 \%\\30.36 ms\\1072.16 ms\end{tabular} & \begin{tabular}[c]{@{}c@{}}98.1 KB\\91.12 \%\\175.90 ms\\1072.16 ms\end{tabular} & \begin{tabular}[c]{@{}c@{}}22.5 KB\\89.17 \%\\0.34 ms\\1264.46 ms\end{tabular}           & \begin{tabular}[c]{@{}c@{}}43.0 KB\\89.19 \%\\0.26 ms\\1075.24 ms\end{tabular} & \begin{tabular}[c]{@{}c@{}}23.2 KB\\89.16 \%\\0.37 ms\\1667.54\end{tabular}     \\ 
\hline
$CNN_2$ & \begin{tabular}[c]{@{}l@{}}i. 271.4 KB\\ii. 98.80 \%\\iii. 344.43 ms\\iv. 1255.09 ms \end{tabular}                                        & \begin{tabular}[c]{@{}c@{}}282.8 KB\\97.11 \%\\30.80 ms\\1414.21 ms\end{tabular} & \begin{tabular}[c]{@{}c@{}}98.1 KB\\99.10 \%\\187.84 ms\\1074.45 ms\end{tabular} & \begin{tabular}[c]{@{}c@{}}22.5 KB\\98.03 \%\\0.38 ms\\1092.26 ms\end{tabular}           & \begin{tabular}[c]{@{}c@{}}43.0 KB\\98.04 \%\\0.62 ms\\1024.50 ms\end{tabular} & \begin{tabular}[c]{@{}c@{}}23.2 KB\\98.05 \%\\0.50 ms\\1605.62 ms\end{tabular}  \\
\hline
\end{tabular}
\label{pre_and_post}
\end{table*}

\fakeparagraph{Number of threads} Limit the number of threads initialized by NNs for computation.

\fakeparagraph{Low-level optimization} Perform low-level optimization of convolution operations using the method presented  ~\cite{liu2019optimizing}. This method adds flexibility in searching for the best implementation of a specific convolution workload on a particular architecture and allows us to optimize the whole computation graph by choosing proper data layouts between operations to eliminate unnecessary data layout transformation overheads.

\fakeparagraph{Linear algebraic properties} Analyze the linear algebraic properties~\cite{cong2014minimizing} of a NN model and apply algorithms such as Strassen Gaussian elimination, Winograd's minimal ﬁltering \cite{strassen1969gaussian} to reduce the computational workload, resulting in increased available memory.

\subsection{Kernels Optimization} \label{kernelopt}

The general C/C++ implemented reference kernels for MCUs (presented in Fig. \ref{aiot_boards}) need platform-specific hardware optimizations. For example, libraries such as NNPack~\cite{nnpack}, which provides manually optimized NN operators on ARM CPUs, cannot optimize kernels of models targeted to be deployed on a wide range of tiny hardware. 

In the following, we present library independent kernel optimization techniques that are generic across a wide range of resource-constrained hardware for guaranteeing no runtime performance bottlenecks.

\begin{itemize}

\item Remove excess modules and components inside the project directory before building a project. This reduces the size of the compiled kernel and also aids the MCUs to boot faster.

\item  Group multiple operators together within a single kernel. Performing this task will improve efficiency due to better memory locality.

\item  Matrix multiplication is a computationally intensive task, yet the main computation kernel that needs to be used during convolution operations. In the context of matrix multiplication on tiny hardware, we recommend using the LIBXSMM~\cite{heinecke2016libxsmm} to improve the kernel performance because it goes deep into the assembly code level for improving small matrix multiplication tasks. Also, we recommend implementing the matrix multiplication kernel with 2x2 kernels in order to save on the total number of load instructions while also enabling some data re-usage.

\item The computationally intense convolutions traverse their operands many times during computation. Hence, managing the layout of the data fed to NNs is critical for reducing memory access overheads. We also recommend applying algorithms to optimize convolution in a single thread (thread optimization) to reduce memory access overheads.

\item The convolutions should be partitioned into disjoint pieces to achieve parallelism. At the CPU level, off-the-shelf multithreading solutions such as OpenMP~\cite{openMP} (used by the Intel MKL-DNN kernel library) are used to achieve parallelism via shared memory multiprocessing. But for MCUs, such approaches are not suited. For MCUs, self-customized thread pooling techniques should be used to reduce overheads while launching and suppressing threads, to reduce performance jitters while adding threads. Using such a self-customized thread pool provides full control of the IoT application while maintaining performance across different MCU platforms.

\end{itemize}

\begin{table*}
\small\addtolength{\tabcolsep}{-2pt}
\centering
\caption{Post-training optimization of the pre-training optimized model versions: a. Comparing pruned models with its post-training optimized versions. b. Comparing quantization-aware trained models with its post-training optimized versions.}
\label{joint_model_opt}
\begin{tabular}{|c|l|c|c|c|c|c|c|} 
\hline
         & \multicolumn{1}{c|}{\begin{tabular}[c]{@{}c@{}}Pruned\\ model \end{tabular}}                     & \begin{tabular}[c]{@{}c@{}}a. Pruned + \\ Int with float \\ fallback \\ quantized \\ version \end{tabular} & \begin{tabular}[c]{@{}c@{}}a. Pruned +\\ Float16 \\ quantized \\ version \end{tabular} & \begin{tabular}[c]{@{}c@{}}a. Pruned +\\ Int only\\ quantized \\ version \end{tabular} & \begin{tabular}[c]{@{}c@{}}Q-aware\\ trained model \end{tabular}                 & \begin{tabular}[c]{@{}c@{}}b. Q-aware + \\ Int with float \\ fallback \\ quantized \\ version \end{tabular} & \begin{tabular}[c]{@{}c@{}}b. Q-aware +\\ Float16 \\ quantized\\ version \end{tabular}  \\ 
\hline
$CNN_1$  & \begin{tabular}[c]{@{}l@{}}i. 98.1 KB\\ii. 91.12 \%\\iii. 175.90 ms\\iv. 1072.16 ms\end{tabular} & \begin{tabular}[c]{@{}c@{}}22.5 KB\\88.81 \%\\0.34 ms\\1722.6 ms\end{tabular}                              & \begin{tabular}[c]{@{}c@{}}43.0 KB\\88.87 \%\\0.25 ms\\1063.4 ms\end{tabular}          & \begin{tabular}[c]{@{}c@{}}23.2 KB\\88.81 \%\\0.33 ms\\1738.23 ms\end{tabular}         & \begin{tabular}[c]{@{}c@{}}108.7 KB\\88.32 \%\\30.36 ms\\1072.16 ms\end{tabular} & \begin{tabular}[c]{@{}c@{}}24.1 KB\\88.26 \%\\0.36 ms\\2246.4 ms\end{tabular}                               & \begin{tabular}[c]{@{}c@{}}43.6 KB\\88.29 \%\\0.28 ms\\1555.74 ms\end{tabular}          \\ 
\hline
$CNN_2$  & \begin{tabular}[c]{@{}l@{}}i. 98.1 KB\\ii. 99.10 \%\\iii. 187.84 ms\\iv. 1074.45 ms\end{tabular} & \begin{tabular}[c]{@{}c@{}}22.5 KB\\97.85 \%\\0.25 ms\\1095.82 ms\end{tabular}                             & \begin{tabular}[c]{@{}c@{}}43.0 KB\\97.84 \%\\0.35 \%\\1074.45 ms\end{tabular}         & \begin{tabular}[c]{@{}c@{}}23.2 KB\\97.85 \%\\0.38 ms\\1606.31 ms\end{tabular}         & \begin{tabular}[c]{@{}c@{}}282.8 KB\\97.11 \%\\30.80 ms\\1414.21 ms\end{tabular} & \begin{tabular}[c]{@{}c@{}}24.1 KB\\97.10 \%\\0.51 ms\\2127.31~ms\end{tabular}                              & \begin{tabular}[c]{@{}c@{}}43.6 KB\\97.11 \%\\0.30 ms\\1479.34 ms\\\end{tabular}        \\
\hline
\end{tabular}
\label{joint_model_opt}
\end{table*}

When users aim to perform an advanced level of optimization, we recommend optimizing the matrix multiplications, CONVs and pooling layers, and activation functions, as explained in the rest of this section.

\fakeparagraph{Optimized Matrix \& Matrix Vector Multiplication Kernels} If MCUs contain a large number of registers, we can implement large kernels to obtain superior multiplication performance. But in reality, for example, the latest Arm Cortex-M (32-bit RISC processor cores) has only 16 architectural registers, including Link \& Program Counter registers. So, we use only 2x2 kernels for reducing operands loading cost into registers. In our CNN, the batch size of the fully-connected layers is one, so we perform matrix-vector multiplication (we view vector as one column matrix) using a 2x1 kernel for speedups. Since storage is expensive in MCUs, we quantized the weights (see Section \ref{PTQ}) to reduce its size. In all the cases, since these quantized weights are stored and re-used during onboard inference, we found that reordering the matrix weights can reduce the pointer accesses. We recommend users to inherit the weight interleaving method from \cite{lai_2018_cmsisnn} to implement the weights reordering tasks.

\fakeparagraph{CONVs and Pooling Optimization for Efficiency} Our CNN converts the MNIST's 28x28 image's pixels into a matrix, an essential step, but consumes memory to store pixels and output matrix. So, depending on the size (2x2) of previously explained matrix-multiplication kernels, the optimized convolution kernel limits its extension/spreading over the image pixels, thus resulting in reduced memory consumption. Pooling (downsamples data in matrix generated by CONVs) is one of the memory-bound operations linked with CONVs. Currently, a nested for-loop approach is used to iterate throughout the window, like in the Caffe framework. But to improve efficiency, in limited memory footprint, we recommend \textit{splitting the pooling operation into two parts} (for both average and max pooling), namely width and height pooling, so the operations to find the maximum or average is the same for both axes, resulting in reduced total operations.

\fakeparagraph{ReLU Activation Function Optimization} ReLU, sigmoid, and tanh are the commonly used activation functions to add non-linearities in the network. In a regular high-resource setting, we generally use the default ReLU layer from TensorFlow. But when users are aiming to deploy and execute their models on low-resource hardware like the AIoT boards, we recommend replacing the default ReLU with its optimized version from \cite{lai_2018_cmsisnn}. Their optimized Single instruction, multiple data Within a Register (SWAR) based ReLU claims to achieve a 4x speed-up for most cases.

\subsection{Deployment of Optimized Models} \label{ModelonMCU}

In this section, using Fig. \ref{fig:rce-nn_steps}, we outline the necessary steps from our recent Resource Constrained Edge - Neural Networks~(RCE-NN) pipeline \cite{BharathRCENN}, to deploy and execute optimized models on MCU-based devices, shown in Fig. \ref{aiot_boards}.

\subsubsection{Model to FlatBuffer Conversion}

It converts models into FlatBuffer, using FlatBuffer's cross-platform serialization library or by also using the TF Lite converter. After conversion, the resulting FlatBuffer format file contains direct data structures of the trained NN. This data structure contains information arrays with a graph consisting of subgraphs, where each subgraph consists of a list of tensors and operators. After this stage, since the flat buffers of the NNs for IoT use-cases are memory-mapped, they can be utilized directly from disk/flash without any loading or parsing tasks, and with zero additional memory requirements for accessing the data (the only memory required to access data is that of the buffer).

\subsubsection{Model Translation}

Most MCUs in edge devices do not have native filesystem support, hence we convert the quantized version of the trained model into a C array, and compile it along with the program for the IoT application which is to be executed on the edge device. In our pipeline, to perform this conversion we use a UNIX command, which generates the C source file containing the quantized model as a char array. 

\subsubsection{Application Integration and Deployment}

\begin{table*}
\centering
\small\addtolength{\tabcolsep}{-3.5pt}
\caption{Comparing original models with its: a. Operations optimized version, b. Post-training optimized versions of the operations optimized model, c. Graph optimized version, and d. Post-training optimized versions of the graph optimized model.}
\label{operations_opt_then_quant}
\begin{tabular}{|c|l|c|c|c|c|c|c|c|c|} 
\hline
\multicolumn{2}{|c|}{\begin{tabular}[c]{@{}c@{}}Original\\models \end{tabular}}                              & \begin{tabular}[c]{@{}c@{}}a. Operations\\ (Ops)\\ Optimized\\ version \end{tabular} & \begin{tabular}[c]{@{}c@{}}b. Ops Opt \\ + Int with \\ float fallback\\ quantized \\ version \end{tabular} & \begin{tabular}[c]{@{}c@{}}b. Ops Opt \\ + Float16 \\ quantized\\ version \end{tabular} & \begin{tabular}[c]{@{}c@{}}b. Ops Opt \\ + Int only\\ quantized\\ version \end{tabular} & \begin{tabular}[c]{@{}c@{}}c. Graph\\ Optimized\\ version \end{tabular}         & \begin{tabular}[c]{@{}c@{}}d. Graph Opt \\ + Int with \\ float fallback\\ quantized \\ version \end{tabular} & \begin{tabular}[c]{@{}c@{}}d. Graph \\ Opt + \\ Float16 \\ quantized\\ version \end{tabular} & \begin{tabular}[c]{@{}c@{}}d. Graph \\ Opt + \\ Int only\\ quantized\\ version \end{tabular}  \\ 
\hline
$CNN_1$  & \begin{tabular}[c]{@{}l@{}}i. 271.4 KB\\ii. 89.43 \%\\iii.~326.69 ms\\iv.~1419.27 ms\end{tabular}    & \begin{tabular}[c]{@{}c@{}}273.7 KB\\85.75 \%\\51.3 ms\\1796.4 ms\end{tabular}       & \begin{tabular}[c]{@{}c@{}}22.7 KB\\85.94 \%\\0.12 ms\\763.7 ms\end{tabular}                               & \begin{tabular}[c]{@{}c@{}}43.7 KB\\85.78 \%\\0.06 ms\\713.69 ms\\\end{tabular}         & \begin{tabular}[c]{@{}c@{}}23.8 KB\\86.2 \%\\0.11 ms\\1406.5 ms\\\end{tabular}          & \begin{tabular}[c]{@{}c@{}}98.2 KB\\89.59 \%\\24.9 ms\\1204.7 ms\\\end{tabular} & \begin{tabular}[c]{@{}c@{}}22.5 KB\\89.59 \%\\0.27 ms\\1139.8 ms\end{tabular}                                & \begin{tabular}[c]{@{}c@{}}43.0 KB\\89.59 \%\\0.33 ms\\1103.3 ms\end{tabular}                & \begin{tabular}[c]{@{}c@{}}23.2 KB\\89.70 \%\\0.33\\1763.4 ms\end{tabular}                    \\ 
\hline
$CNN_2$  & \begin{tabular}[c]{@{}l@{}}i. 271.4 KB\\ii. 97.94 \%\\iii.~344.43 ms\\iv.~1255.09 ms\end{tabular} & \begin{tabular}[c]{@{}c@{}}273.7 KB\\96.47 \%\\52.9 ms\\1880.7 ms\end{tabular}       & \begin{tabular}[c]{@{}c@{}}22.7 KB\\96.46 \%\\0.07 ms\\757.82 ms\\\end{tabular}                            & \begin{tabular}[c]{@{}c@{}}43.7 KB\\96.47 \%\\0.07 ms\\683.9 ms\end{tabular}            & \begin{tabular}[c]{@{}c@{}}23.8 KB\\96.42 \%\\0.12 ms\\1421.5 ms\end{tabular}           & \begin{tabular}[c]{@{}c@{}}98.2 KB\\98.03 \%\\28.3 ms\\1194.0 ms\end{tabular}   & \begin{tabular}[c]{@{}c@{}}22.5 KB\\98.05 \%\\0.28 ms\\1210.9 ms\end{tabular}                                & \begin{tabular}[c]{@{}c@{}}43.0 KB\\98.03 \%\\0.34 ms\\1079.5 ms\\\end{tabular}              & \begin{tabular}[c]{@{}c@{}}23.2 KB~\\98.07 \%\\0.29 ms\\1724.2 ms~\end{tabular}               \\
\hline
\end{tabular}
\label{operations_opt_then_quant}
\end{table*}

This method fuses the \emph{c-byte array} of NNs with the main program for an IoT use-case.Finally, the method from this last step should be used to flash the binaries of a NN model on the MCU-based hardware devices.

\section{Evaluation}\label{sec:evaluation}

We selected to perform experiments on CNNs since it is a popular subclass of NNs that is widely used to solve various problems in domains such as image classification, pose estimation, semantic segmentation, text detection, etc. For experiments, we use the standard MNIST Fashion (produces $CNN_1$) and MNIST Digits (produces $CNN_2$) datasets to train a basic CNN whose architecture is shown in Fig. \ref{model_comp}. a. Both these datasets are imported via the \textit{tf.keras.dataset.name} function with its default train and test sets. After importing, we apply all suitable optimizers before, during, and after training CNNs and report the memory conservation, accuracy, and inference speedups achieved after realizing each optimizer component. During experiments, for statistical validation, the reported inference time corresponds to the average of 5 runs.

\fakeparagraph{Evaluating Pre-training Optimization Techniques} We first apply the pruning technique ~(Section~\ref{pretrain}) on CNNs and present its evaluation results in Table~\ref{pre_and_post}. a and the changes in inference time and size in Fig. \ref{model_comp}. c. We also perform quantization-aware training of CNNs and show the changes in Fig. \ref{model_comp}. b and its evaluation results in Table \ref{pre_and_post}. a.

\fakeparagraph{Evaluating Post-training Optimization Techniques} As explained in Section \ref{PTQ}, we quantized the original CNN's Float32 weights and activations to Float16 values. Users can use this Float16 quantization when they want to achieve reasonable compression rates (we obtain approx. 6x compression), without loss of precision (we experience only 0.01 \% loss in accuracy). Also, Float16 models run on small CPUs without modification. In Fig. \ref{model_comp}. e, we show the Float16 quantized model's architecture, inference time, and size changes. Next, we performed Int with float fallback quantization on original CNNs and show its architecture and performance in Fig. \ref{model_comp}. d, next to other quantization results, that can be compared with the original model in Fig. \ref{model_comp}. a. We present the evaluation results of all the thus produced post-training quantized models in Table \ref{pre_and_post}. b. We realized this method to convert our CNN's weights \& activation to 8-bit integers and show its architecture, inference time, and size changes in Fig. \ref{model_comp}. f. Here, the size reduced and inference time improved since, after quantization, the inference is carried out using integer-only arithmetic.

Next, as explained in the \textit{joint pre and post-training model optimization} part of Section \ref{PTQ}, we performed post-training optimization of the pre-training optimized model and present the evaluation results of resultant CNNs in Table \ref{joint_model_opt}.

\fakeparagraph{Evaluating Graph Optimization Components} We implemented and performed all applicable arithmetic simplification rewrites and graph structure optimization tasks from Section \ref{GO}, on CNNs and present their evaluation results in Table \ref{operations_opt_then_quant}. c. In Fig. \ref{model_comp}. h, we show the graph optimized model's inference time and size changes. Next, as explained in the \textit{joint graph and post-training model optimization} part of Section \ref{GO}, we performed post-training optimization of the graph optimized CNNs and present the results in Table \ref{operations_opt_then_quant}. d.

\fakeparagraph{Evaluating Operations Optimization Components} We implemented and applied the explained technique from Section \ref{OO} on CNNs and present their evaluation results in Table \ref{operations_opt_then_quant}. a. next to the results of the original CNNs. In Fig. \ref{model_comp}. g, we show the operations optimized model's architecture, inference time, and size changes. Next, we also performed the \textit{joint operations and post-training model optimization} part from Section \ref{OO} and present the evaluation results for CNNs in Table \ref{operations_opt_then_quant}. b.

\fakeparagraph{Evaluating Model Deployment}
We take the \emph{joint graph and post-training optimized} CNNs~(both $CNN_1$ and $CNN_2$) from Section~\ref{PTQ}, then using model deployment techniques~(Section~\ref{ModelonMCU})  we load and execute them on the ESP32 and nrf52840 boards shown in Fig. \ref{aiot_boards}. We report that the accuracy obtained by executing CNNs on both boards is the same until its first decimal point: 89.5\% for $CNN_1$, 98.03\% for $CNN_2$ on both boards. From this, it is clear that when the proposed model deployment technique is used to deploy and execute the CNNs optimized using our multi-component sequence, they perform the same, irrespective of the target hardware.

\subsection{Results Analysis} \label{result_analysis}

In this section, we perform analysis based on the experiment results from Table \ref{pre_and_post} - \ref{operations_opt_then_quant}.

\fakeparagraph{Best Optimization Sequence for Smallest Model Size} When users want the smallest possible trained model, we recommend performing \emph{Joint graph and post-training model optimization} from Section \ref{GO}. Since, it is apparent from Table \ref{operations_opt_then_quant}. d, that this \emph{Graph optimized then integer with float fallback quantized version} of the original CNN has the smallest model size of 22.5 KB (12.06 x times smaller than original CNN). Although the Pruned then int with float fallback quantized model version has the same size, the accuracy after optimization drops by 2.31 \% (see Table \ref{joint_model_opt}. a), whereas the former sequence has no accuracy drop.
            
\fakeparagraph{Best Optimization Sequence for Accuracy Preservation} When the target hardware can accommodate a few extra KB, naturally we would try to fit the top-performing model. In such cases, we recommend to load and use the \emph{Graph optimized then integer only quantized version} since training then optimizing using this sequence preserved accuracy for both the datasets and in fact, for MNIST Fashion, the accuracy increased by 0.27 \%, and by 0.13 \% for MNIST Digits (see table \ref{operations_opt_then_quant}. d).

\fakeparagraph{Best Optimization Sequence for Fast Inference} For real-time applications, we naturally tend to load and use the fastest inference results producing models. In such cases, we recommend the \emph {Joint operations and post-training optimization} from Section \ref{OO}, since the \emph{Operations optimized then float16 quantized version} (see Table \ref{operations_opt_then_quant}. b) produces the fastest unit inference results in 0.06 ms (orders of magnitude faster than original CNN).

We show how to apply our optimizer components on popular pre-trained CNNs. We take the Mobilenet v2, Inception v3, Resnet v2, which are 14 MB, 95 MB, 178 MB each, and apply just the post-training model optimization component from Section \ref{PTQ}. The resultant models are 3.5 MB, 23 MB, 44 MB, i.e., 4x, 4.13x, 4.04x times smaller than their respective original modes. Depending on the user's goal, we let them explore other presented joint optimization sequences that can make the above CNN-based models much smaller or faster or show the top accuracy.

\section{Conclusion} \label{concl}

We presented our multi-component model optimizer, a sequence that researchers and developers can follow to optimize various CNNs for making it executable on multiple resource-constrained IoT hardware like MCUs, small CPUs and AIoT boards. Our optimization sequence can be applied to the models from a growing number of use-cases such as anomaly detection, predictive maintenance, robotics, voice recognition, machine vision, etc., to enable their standalone execution on the boundaries of the IoT architecture. By open-sourcing the implementation of our optimizer components, we believe the transparent design to open future avenues for a broad-spectrum of applied research works and also interconnect the high-performance computing community (studies with algorithms that get merged with ML frameworks like TensorFlow) with the TinyML community (studies that design resource-friendly models for embedded systems).

In future work, we plan to evaluate the optimization efficiency of workload and kernel optimization components. We also plan to apply our optimization sequence on advanced models like Tiny-YOLO (9 convolutional layers), SqueezeNet (5MB just for parameters), etc. Such models with more layers and parameters benefit the most from our multi-component optimization sequence, so can outperform the state-of-the-art methods that aim to reduce the model size and improve inference speed. We also plan to execute models optimized using our sequence on IoT hardware and benchmark the real-life memory conservation and inference speedups.


\balance \bibliographystyle{IEEEtran}
\bibliography{bib.bib}

\newpage

\begin{IEEEbiography}[{\includegraphics[width=1in,height=1.25in,clip,keepaspectratio]{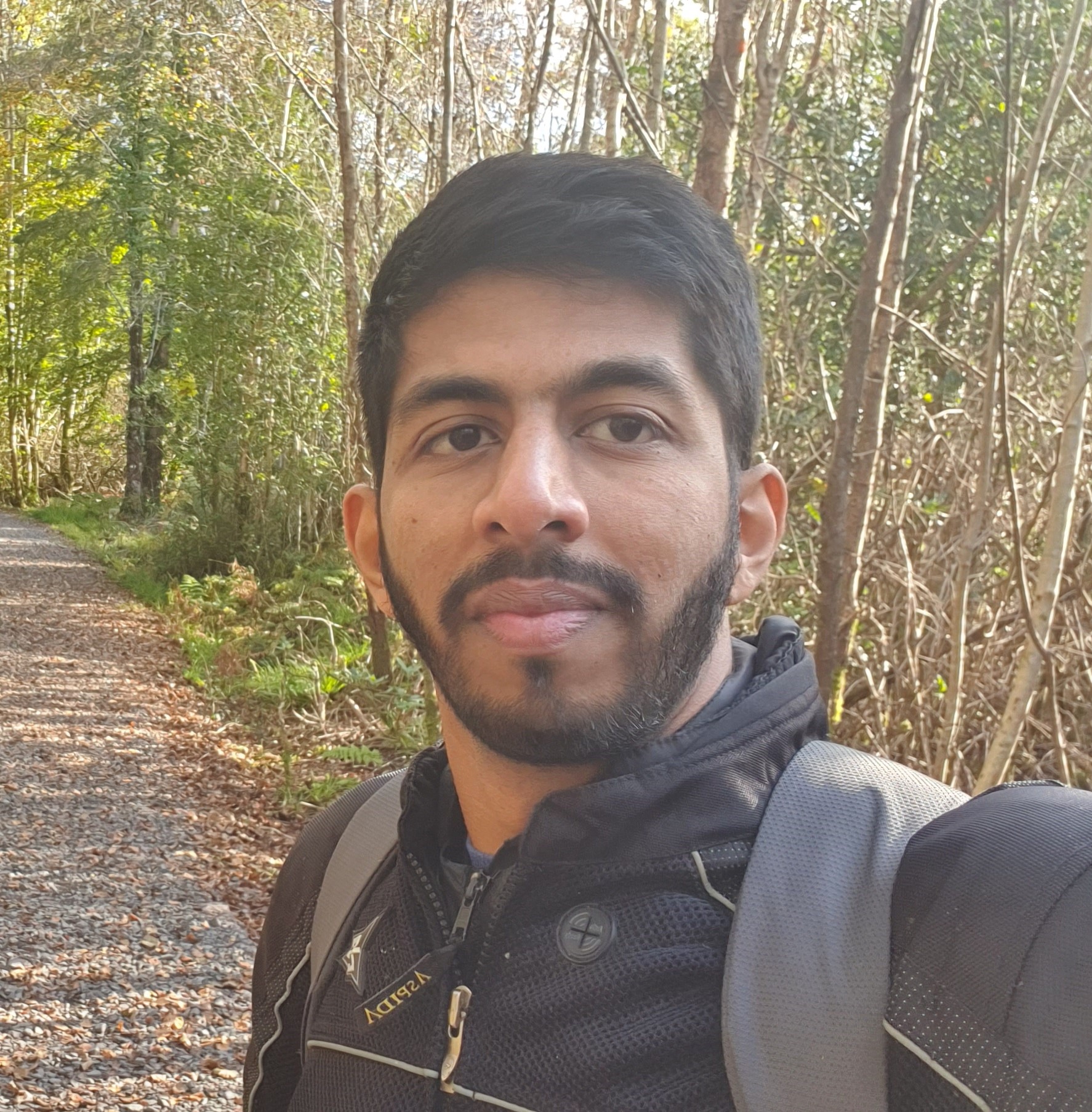}}]{Bharath Sudharsan} is a Senior AI ML Software Engineer at the Crowd Intelligence Team, General Motors, Ireland. He did his Ph.D. at the Data Science Institute, NUI Galway, with funding from CONFIRM SFI Research Centre for Smart Manufacturing. During  Ph.D., he contributed to science by publishing 15+ first-author full papers in top-tier venues such as IEEE Internet Computing, IEEE IoT Journal, ECML PKDD, ACM IoT, IEEE SCC, IEEE UIC, IEEE BigData, and also provided 10+ demos. He obtained MEngg from NUI Galway in Electronics and Computer Engineering.
\\ Homepage: \url{https://bharathsudharsan.github.io/profile/}
\end{IEEEbiography}

\begin{IEEEbiography}[{\includegraphics[width=1in,height=1.25in,clip,keepaspectratio]{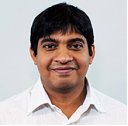}}]{Pankesh Patel} Before joining Artificial Intelligence Institute, University of South Carolina, Dr. Pankesh Patel was Technology Consultant at Jupyter. He was hired at Jupyter, to develop AI and Cloud-based Intelligent Doorbell products for the Australian market. Before joining these positions, he was a Senior Research Scientist at Fraunhofer USA and a Research Scientist in Industrial Software System (ISS) group at ABB Corporate Research-India. Both at Fraunhofer USA and ABB, he focused on the implementation of Industry 4.0 techniques and methodologies in commercial environments. His academic background and research work focus on building software development tools to easily develop applications in the cross-section of the Internet of Things/Industry 4.0, Artificial Intelligence, Edge, and Cloud Computing. 

He is a winner of the prestigious Marie-Curie fellowship at SFI Confirm Centre for Smart Manufacturing, Data Science Institute, NUIG Galway, Ireland. In the past 7 years, he has published 40+ publications research articles in prestigious conferences and delivered several talks as a keynote and invited speaker. He finished his Ph.D. in Computer Science (with a Très Honorable award) from the University of Paris VI (UPMC), France. His Ph.D. was funded by the French National Institute for Research in Computer Science and Control (INRIA) Paris, France. 

\end{IEEEbiography}

\begin{IEEEbiography}[{\includegraphics[width=1in,height=1.25in,clip,keepaspectratio]{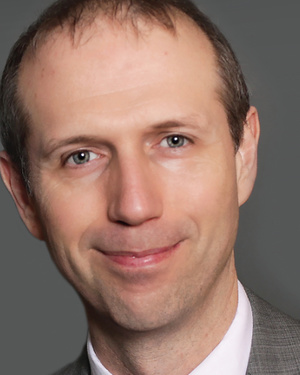}}]{JOHN BRESLIN} is a Professor (Personal Chair) in Electronic Engineering at NUI Galway, where he is Director of the TechInnovate/AgInnovate programmes. John has taught electronic engineering, computer science, innovation, and entrepreneurship topics during the past two decades. Associated with three SFI Research Centres, he is a Co-Principal Investigator at Confirm (Smart Manufacturing) and Insight (Data Analytics), and a Funded Investigator at VistaMilk (AgTech). He has written 200+ peer-reviewed academic publications (h-index of 42, 7400 citations, best paper awards from IoT, DL4KGS, SEMANTiCS, ICEGOV, ESWC, PELS), and co-authored the books "Old Ireland in Colour", "The Social Semantic Web" and "Social Semantic Web Mining". He co-created the SIOC framework (Wikipedia article), implemented in hundreds of applications (by Yahoo, Boeing, Vodafone, etc.) on at least 65,000 websites with 35 million data instances.
\\Homepage: \url{http://www.johnbreslin.com/}
\end{IEEEbiography}

\begin{IEEEbiography}[{\includegraphics[width=1in,height=1.25in,clip,keepaspectratio]{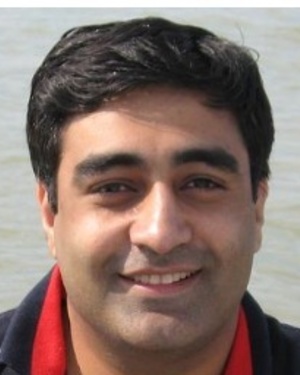}}]{MUHAMMAD INTIZAR ALI} is an Assistant Professor in the School of Electronic Engineering, Dublin City University. He received the PhD (Hons) degree from the Vienna University of Technology, Austria, in 2011. His research interests include semantic Web, data analytics, Internet of Things (IoT), linked data, federated query processing, stream query processing, and optimal query processing over large scale distributed data sources. He is actively involved in various EU funded and industry-funded projects aimed at providing IoT enabled adaptive intelligence for smart applications. He serves as a PC Member of various journals, international conferences, and workshops. 
\\Homepage: \url{http://www.intizarali.org/}
\end{IEEEbiography}

\begin{IEEEbiography}[{\includegraphics[width=1in,height=1.25in,clip,keepaspectratio]{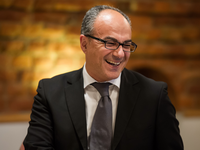}}]{Schahram Dustdar} is Full Professor of Computer Science heading the Research Division of Distributed Systems at the TU Wien, Austria. He holds several honorary positions: Francqui Chair Professor at University of Namur, Belgium (2021-2022), University of California (USC) Los Angeles; Monash University in Melbourne, Shanghai University, Macquarie University in Sydney, University Pompeu Fabra, Barcelona, Spain. From Dec 2016 until Jan 2017 he was a Visiting Professor at the University of Sevilla, Spain and from January until June 2017 he was a Visiting Professor at UC Berkeley, USA.

He is founding co-Editor-in-Chief of ACM Transactions on Internet of Things (ACM TIoT) as well as Editor-in-Chief of Computing (Springer). He is an Associate Editor of IEEE Transactions on Services Computing, IEEE Transactions on Cloud Computing, ACM Computing Surveys, ACM Transactions on the Web, and ACM Transactions on Internet Technology, as well as on the editorial board of IEEE Internet Computing and IEEE Computer. Dustdar is recipient of multiple awards: TCI Distinguished Service Award (2021), IEEE TCSVC Outstanding Leadership Award (2018), IEEE TCSC Award for Excellence in Scalable Computing (2019), ACM Distinguished Scientist (2009), ACM Distinguished Speaker (2021), IBM Faculty Award (2012). He is an elected member of the Academia Europaea: The Academy of Europe, where he is chairman of the Informatics Section, as well as an IEEE Fellow (2016), an Asia-Pacific Artificial Intelligence Association (AAIA) President (2021) and Fellow (2021) and a Member of the Academy of the United Nations Sciences and Technology Organization (2021). 
\\Homepage: \url{https://dsg.tuwien.ac.at/team/sd/}
\end{IEEEbiography}

\begin{IEEEbiography}[{\includegraphics[width=1in,height=1.25in,clip,keepaspectratio]{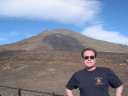}}]{Albert Y. Zomaya} 

Albert Y. Zomaya is currently the Chair Professor of High Performance Computing \& Networking in the School of Computer Science, University of Sydney. He is also the Director of the Centre for Distributed and High Performance Computing which was established in late 2009. Professor Zomaya published more than 550 scientific papers and articles and is author, co-author or editor of more than 20 books. He served as the Editor in Chief of the IEEE Transactions on Computers (2011-2014). Currently, Professor Zomaya serves as a Founding Editor in Chief of the IEEE Transactions on Sustainable Computing, Founding Co-Editor in Chief of the IET Cyber-Physical Systems, and Associate Editor-in-Chief (Special Issues), Journal of Parallel and Distributed Computing. He also serves as associate editor for 22 leading journals, such as, the ACM Computing Surveys, ACM Transactions on Internet Technology, and IEEE Transactions on Cloud Computing. Professor Zomaya is the Founding Editor of several book series, such as, the Wiley Book Series on Parallel and Distributed Computing, Springer Scalable Computing and Communications, and the IET Book Series on Big Data.

Professor Zomaya has delivered more than 180 keynote addresses, invited seminars, and media briefings and has been actively involved, in a variety of capacities, in the organization of more than 700 conferences.
\end{IEEEbiography}

\begin{IEEEbiography}[{\includegraphics[width=1in,height=1.25in,clip,keepaspectratio]{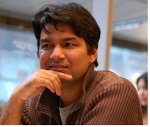}}]{Rajiv Ranjan} is an Australian-British computer scientist, of Indian origin, known for his research in Distributed Systems (Cloud Computing, Big Data, and the Internet of Things). He is University Chair Professor for the Internet of Things research in the School of Computing of Newcastle University, United Kingdom.  He is the director of Networked and Ubiquitous Systems Engineering (NUSE) Group, jointly with Dr. Graham Morgan, in the School of Computing. He is also the Academic Director of School of Computing and the Research Director of Newcastle Urban Observatory. He is an internationally established scientist in the area of Distributed Systems (having published over 250 scientific papers out of which about 50 papers in the IEEE/ACM Transactions Journals). He is a fellow of Academia Europaea. He has secured more than \$32 Million+ AUD (£16 Million+ GBP, with collaborators) in the form of competitive research grants from both public and private agencies.

Prof. Ranjan is thankful to the research community for recognising his research through citations and using the open-source tools that he help develop along with collaborators. He is ranked by Microsoft Academic as one of the Top Authors in Cloud Computing (2010-2020), Big Data (1997-2021), Quality of Service (2000-2019), Resource Management (2000-2019), and Services Computing (1999-2018).  According to recent (2020) bibliometric study by the Stanford University (https://bit.ly/3ndOXlN), he is one of the highly cited authors in distributed computing field.  Other bibliometric evidences (Guide2Research and UCLA) also reveal immense support and acceptance from his peers. 
\\Homepage: \url{https://rajivranjan.net/}
\end{IEEEbiography}

\EOD

\end{document}